\documentclass[10pt]{article}

\usepackage[english]{babel} 
\usepackage{custom_commands}
\usepackage{custom_aliases}
\usepackage{subcaption}
\usepackage{graphicx}
\usepackage{amsmath}
\usepackage{authblk}
\usepackage[normalem]{ulem}
\usepackage{algpseudocode}
\usepackage{algorithm}
\newcommand{\Rea}{{\rm I\!R}}
\usepackage{bm}
\usepackage{url}
\usepackage[legalpaper, margin=3cm]{geometry}

\begin{document}

\title{Wind Farm Layout Optimisation using Set Based Multi-objective Bayesian Optimisation}

\author[1]{Tinkle Chugh \thanks{t.chugh@exeter.ac.uk}}
\author[1]{Endi Ymeraj \thanks{ey238@exeter.ac.uk}}

\affil[1]{University of Exeter, Exeter EX4 4QD, UK}

\date{}
\maketitle

\begin{abstract}
Wind energy is one of the cleanest renewable electricity sources and can help in addressing the challenge of climate change. One of the drawbacks of wind-generated energy is the large space necessary to install a wind farm; this arises from the fact that placing wind turbines in a limited area would hinder their productivity and therefore not be economically convenient. This naturally leads to an optimisation problem, which has three specific challenges: (1) multiple conflicting objectives (2) computationally expensive simulation models and (3) optimisation over design sets instead of design vectors. The first and second challenges can be addressed by using surrogate-assisted e.g.\ Bayesian multi-objective optimisation. However, the traditional Bayesian optimisation cannot be applied as the optimisation function in the problem relies on design sets instead of design vectors. This paper extends the applicability of Bayesian multi-objective optimisation to set based optimisation for solving the wind farm layout problem. We use a set-based kernel in Gaussian process to quantify the correlation between wind farms (with a different number of turbines). The results on the given data set of wind energy and direction clearly show the potential of using set-based Bayesian multi-objective optimisation.

\end{abstract}


\maketitle

\section{Introduction}
Climate change has become a primary concern that needs to be promptly addressed. The continuously rising temperatures across the globe are causing an increase in the frequency of extreme weather conditions, such as heavy rainfall, which have lead to changing habitat ranges for plants and animals \cite{temp}. It is understood that greenhouse gas emissions are the primary reason for the rise in temperatures, for example, we can see that the electricity sector contributed to 27\% of the total U.S. emissions in 2018 \cite{emission}. Of major concern is also the fact that due to socioeconomic developments and a warmer globe the energy required by all populations across the world is likely to increase \cite{socioeco}.

Renewable energies use natural flows that currently exist in our environment and produce minimal waste which makes them a suitable alternative to fossil fuel based energy. With the desire to implement its commitments under the Paris Agreement \cite{paris}, the European Union has proposed a new set of targets for 2030. Among these targets, there is the desire to reach at least a 32\% share of renewable energy. This is a clear signal that clean energy will receive more attention in the upcoming years.

One of the renewable energies that has increased in popularity is wind energy. This is due to the fact that wind energy is amongst the cleanest source of electricity given the very low greenhouse gas emission and low water consumption \cite{evans}. One of the drawbacks of wind-generated energy is the large space necessary to install a wind farm; this arises from the fact that placing wind turbines in a limited area would hinder their individual productivity and therefore not be economically convenient. Because of this, the optimisation of the layout plays an important role in energy yield. The complexity of this optimisation problem is given by the computational power required to reproduce accurate simulations of the wind across a wind farm, of major complexity is the computation of the wind behind the turbine usually referred to as the wake. Furthermore, it can be difficult to account for the uncertainty in the wind speed and direction. 

In addition to the computational cost and uncertainty in wind speed and directions, the objective function evaluations rely on the design sets instead of design (or decision) vectors. For instance, maximising power as an objective function is an output of the whole wind farm and not of one turbine. Therefore, to design an optimal wind farm, we need to consider the correlations between wind farms, which can have different number of turbines. In other words, the sets can have different number of design vectors. In this paper, we use a set-based kernel in Gaussian process model \cite{gpoversets} and embed it in the Bayesian multi-objective optimisation framework. To the best of our knowledge, set-based multi-objective Bayesian optimisation has never been used to solve a wind farm layout optimisation problem. To be summarised, the contributions of the paper are:
\begin{enumerate}
    \item Handling uncertainty in wind speed and direction by building a probabilistic model.
    \item Handling computationally expensive multiple conflicting objectives by using Bayesian multi-objective optimisation.
    \item Handling correlations between different wind farms (as different design sets) by using a set based kernel.
\end{enumerate}

The rest of the article is structured as follows. In Section 2, we provide an overview of the wind farm layout problem and define the optimisation problem. In Section 3, we explain the Bayesian multi-objective optimisation framework with Gaussian over sets as surrogate models. In Section 4, we provide the results and finally we conclude and provide future research directions in Section 5.




\color{red}

\color{black}

\section{Wind farm layout Optimisation}
\subsection{Background}
The literature behind the wind farm layout optimisation problem is extremely vast. Some of the early attempts to solve the problem date back to 1994, where Mosetti \cite{Mosetti1994} used a genetic algorithm to optimise the layout. In their work, the wake model used was limited to three types of wind cases: single direction, constant intensity with variable direction, and variable intensity with variable direction. 
To date, there have been numerous papers that have improved on Mosetti's work, since then we have seen the development of more accurate wake models which were not limited to specific wind scenarios. Some of these improvements include a Gaussian wake model, Parada et al. \cite{wakegauss}, which resulted in notable improvements only when tested under constant wind scenarios.


Furthermore, multi-objective optimisation including NSGA-II with a simplified wake model has also been used in wind farm layout optimisation \cite{Mittal}. One more rigorous attempt at the wind farm layout optimisation problems is based on computational fluid dynamics (CFD) simulations. These are models that are applied to accurately simulate the wake effect of the turbines often using  Reynolds-averaged Navier–Stokes \cite{RANS} and Large Eddy Simulation \cite{LES} turbulence models. The use of such simulations allows for a better understanding of what the true power output of a wind farm would be. The drawback of the CFD based approach lies in the fact that the evaluations of these models are computationally expensive. Attempts to limit the computational cost of such models include deterministic optimisation (e.g.\ mixed-integer programming \cite{cdfmix}), where an effort was made to reduce the number of CFD simulations required to find optimal solutions. One of the advantages of these mathematical approaches is the fact that they are very robust and can deal with different types of terrains. 


In this paper, we apply a Bayesian model over sets that aim to reduce the computational time to obtain a set of approximated Pareto optimal solutions. For the demonstration of the potential of the approach, we used a simplified Jensen wake model to consider the fluid dynamic aspects. Moreover, we use a joint probability distribution of wind speed and direction with the Jensen model to evaluate the power output.  

\subsection{Wake model}
\begin{figure}[h]
\centerline{\includegraphics[scale=.4]{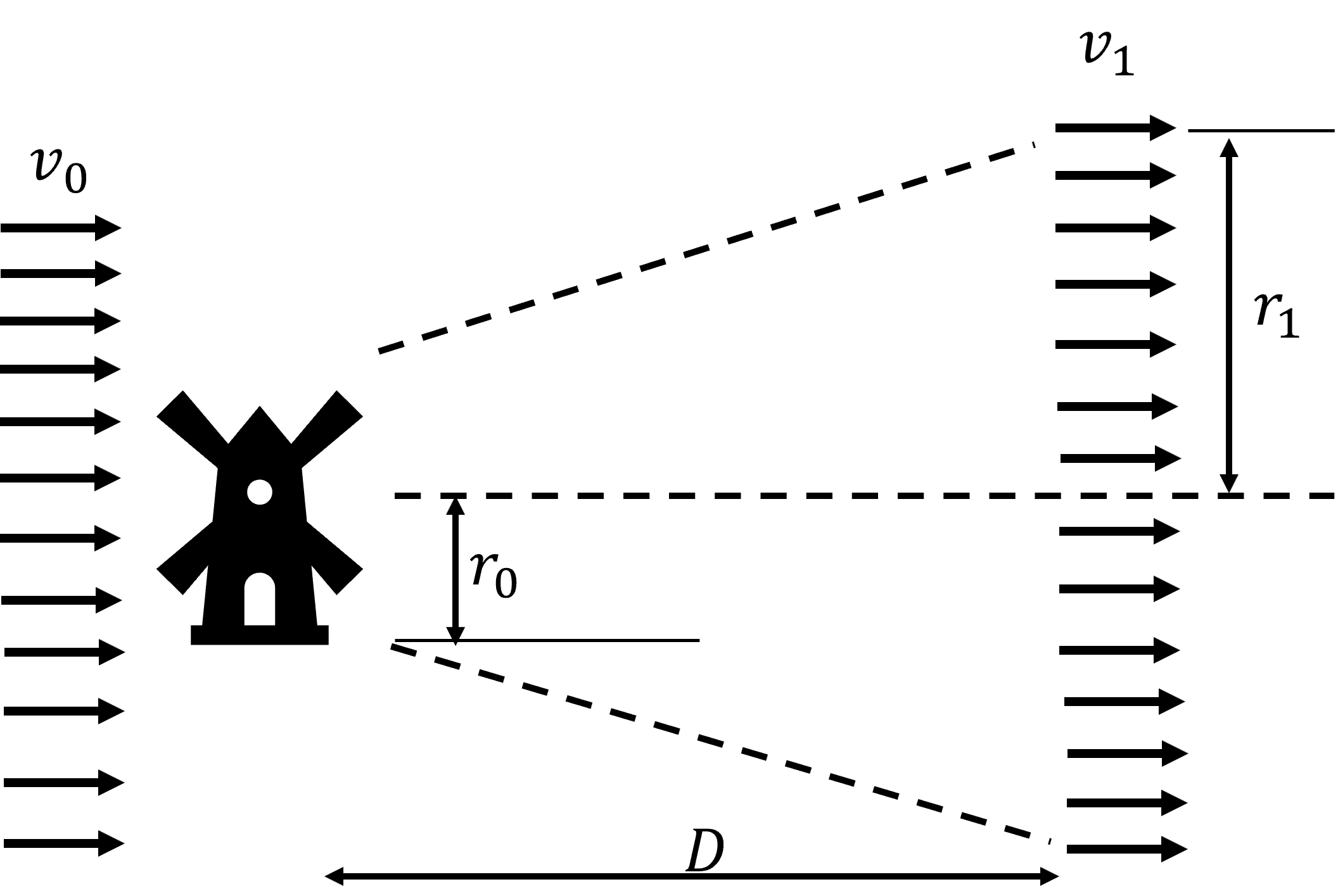}}
\caption{Visual representation of the wake effect from the Jensen's model.}
\label{wake_effect}
\end{figure}

The wake effect considers the wind speeds facing different turbines in a wind farm. The wind speeds for different turbines can be different and depend on the coordinates of turbines and the incoming wind speed (or downstream wind speed) and direction. A simplified well-known wake models is Jensen model \cite{jensen} (also known as park model), which can be used to estimate the wind speed for each turbine. 
This model allows for fast computations of the wake effect of turbines since it assumes that the wake behind the rotor expands linearly and that it is only affected by the distance from the turbine. An example to estimate the wind speed $v_1$ at a distance $D$ from a turbine is shown in Figure \ref{wake_effect}. In the figure, $v_0$, $r_0$ and $r_1$ are downstream wind speed, turbine rotor radius and radius of cone, respectively. The radius of the cone can be estimated using the equation below:
\begin{equation}
    r_1 = r_0 + \alpha D,
\end{equation}
where $\alpha$ is the decay constant and determines the expansion of wake with distance. It is usually calculated with an analytical expression:
\begin{equation}
    \alpha = \frac{0.5}{\log (\frac{z}{z_0})},
\end{equation}
where $z$ is height of the turbine and $z_0$ is the surface roughness of the wind farm. In many cases, the $\alpha$ is kept constant. After applying the conservation of momentum, the reduced wind speed $v_1$ is given by:
\begin{equation}
v_1 = v_0\left[1-\frac{1-\sqrt{1-C_T}}{(1+\alpha \frac{D}{r_1})^2}\right],  
\end{equation}
where $C_T$ is the thrust coefficient. For more details about wake models, see \cite{jensen}.
\color{black}

\subsection{Objective Functions}
One of the main objectives in designing a wind farm is to maximise the power output. In addition, the cost of installing turbines needs to be minimised. The power output depends on the distribution $p(v,\theta)$ of wind speed ($v$) and direction ($\theta$) and the wind speed at different turbines. The distribution $p(v,\theta)$ can be modelled with the historic data of wind speed and direction. In this way, the uncertainty in wind speed and direction can be handled and quantified in estimating the power output. In this work, we used an open source data set provide by Engie \cite{mydata} from 2013-2016. One can use another data set to build a joint distribution of wind speed and direction. 

The input to the wake model is the incoming wind speed, direction and location of wind turbines in the wind farm and the output is the wind speed at different turbines. This wind speed is then used to estimate the power curve using the following equation \cite{kusiak2009line}:
\begin{equation}
\begin{split}
    &P_{curve}(v,\mathbf{x}) = 
    \begin{cases} a\cdot \frac{1+m\cdot\mathrm{exp}({\scriptstyle - v/\tau})}{1+n\cdot\mathrm{exp}({\scriptstyle-v/\tau)}} \:\: if \:\: v \geq v_{cut\_in} \\
    0 \:\:\:\:\:\:\:\:\:\:\:\:\:\:\:\:\:\:\:\:\:\:\:\:\:\:\:\:\:\: elsewhere\end{cases}
\end{split}
\end{equation}
where $(a,m,n,\tau)$ are the parameters (values are taken from \cite{kusiak2009line}) and $v_{cut\_in}$ is the wind speed at which wind turbine becomes productive. The $\mathbf{x}$ represents the coordinate or location of turbines and is an input to the wake model. Once the joint distribution of wind speed and direction and wind speed at different turbines is known, the power output of wind farm is defined as:
\begin{equation}
\hspace{-.3cm}\begin{split}
P_{tot}(X) &= \sum_{x \in X}\sum_{\theta = 0}^{359} \sum_{v = 0}^{v_{max}} P_{curve}(v,\mathbf{x}) p(v,\theta)\\\
P_{tot}(X) &= \mathrm{Expected\:power\:of\:wind\:farm\:X}\\
\theta &= \mathrm{Wind\:direction}\\
v &= \mathrm{Wind\:speed}\\
|X|&=\mathrm{Cardinality\:of\:set\:X\:}\\
    &=\mathrm{(or\:number\:of\:turbines\:in\:the\:wind\:farm)}\\
p(v,\theta) &= \mathrm{Joint\:distribution \:of\:wind\:speed\:and\:direction}\\    
\end{split}    
\label{power_objective}
\end{equation}

As can be seen in the equation above, the expected power of the wind farm depends on the number of turbines. Two wind farms with different number of turbines may give two different output energy. Therefore, to model the objective function, we need to find the correlation between wind farms, which can have different number of turbines. In other words, the correlation must be quantified between design sets (or wind farms) with different number of design vectors (or wind turbines). This work uses the Gaussian process over sets to quantify the correlation between wind farms and embed it into the Bayesian multi-objective optimisation framework. The details are provided in the next section. The second objective is the cost, which can be computed as non-linear function of number of wind turbines \cite{Mosetti1994}:
\begin{equation}
\begin{split}
    \mathrm{cost}(X) = |X| \cdot \left(\frac{2}{3} +\frac{1}{3}\cdot \mathrm{exp}(-0.00174\cdot |X|^2)\right)
\end{split}
\label{cost_objective}
\end{equation}
The cost as the objective function depends only on the number of turbines and does not use computationally expensive wake simulation models.




\section{Multi-objective Bayesian Optimisation over Sets}

In Multi-Objective Bayesian Optimisation, we start with a data set:
$\mathcal{D} = \{(X,Y)| X \in \Rea^{N\times d}, Y \in \Rea^{N\times w} \}$,
where $d$ is the number of variables, $w$ is the number of objectives and $N$ is the number of instances in the data.
With this data, we want to fit a Gaussian Process ($\mathcal{GP}$) model that helps predict objective functions without direct evaluation. When dealing with more than one objective, there are two approaches to build surrogate models. The first consists of building a $\mathcal{GP}$ on a scalarized version of objective functions \cite{scalarizing, Rahat2017, Knowles2006}, with this procedure we are changing the nature of the problem into a single-objective problem; this approach however might not be suited for every type of problem since the relationship between objectives might not always be trivial or easily expressed. The second alternative consists in building a Gaussian Process for each objective\cite{mobopt,Emmerich2006, Emmerich2011, Yang2019}. This latter option is what will be used in this paper. Once we have fitted the surrogate models we aim to use them to search for a new decision vector that maximises an acquisition function. By acquisition function we mean a function that can help us understand if a decision vector is providing a good balance of exploitation and exploration. Various are the options for an acquisition function for example expected improvement \cite{Mockus1975, Jones1998} probability of improvement \cite{prob_of_improvement,Kushner1964} or lower-confidence bound \cite{upper_conf_bound}. In this work, given the surrogates for multiple objectives we will be using the expected hypervolume improvement (EHVI) \cite{EHVI, Emmerich2006, Yang2019,Daulton2020}.

Once the new decision vector ($\mathbf{x}' \in \Rea^{1\times d}$) has been selected after maximising the acquisition function, we compute an expensive evaluation ($\mathbf{y}' = \mathcal{F}(\mathbf{x}'))$, where $\mathcal{F} = (f_1(\mathbf{x}),...,f_w(\mathbf{x}))$ and $\mathbf{y}' \in \Rea^{1\times w}$ and add $(\mathbf{x}',\mathbf{y}')$ to the existing data and repeat the process. The optimisation finishes when a predefined termination criterion is reached (e.g.\ number of expensive evaluations). An outline of the algorithm is shown in Algorithm 1, where $\alpha_{EHVI}$ is the expected hypervolume improvement. 

\begin{algorithm} [t!]
\caption{Multi-objective Bayesian Optimisation}
\label{alg:mbore}
\begin{algorithmic}[]
    \State \textbf{Input}: 
        Data set $\mathcal{D} = \{(X,Y)| X \in \Rea^{N\times d} \enspace  Y \in \Rea^{N\times w} \}$ and $\mathcal{F}(\mathbf{x}) = (f_1(\mathbf{x}),...,f_w(\mathbf{x}))$
     \State \textbf{Output}: Evaluated solutions 
\end{algorithmic}

\begin{algorithmic}[1]
     \While {Termination criterion is not met}
      \State Fit $\mathcal{GP}_i \enspace \forall \enspace  i \in [1,2,...,w]$ 
    \State Find $\mathbf{x'}$ after maximising the acquisition function $\alpha_{EHVI}$ 
  \State Expensively compute $\mathbf{y'} = \mathcal{F}(\mathbf{x'})$ 
  \State Append $\mathbf{x'}, \:\mathbf{y'}$ to $\mathcal{D}$ 
    \EndWhile
\end{algorithmic}
\end{algorithm}


To deal with the type of objective function in the wind layout optimisation problem, the traditional formulation of Gaussian Processes cannot be used. Two are the main reasons for this: Firstly, we have data such that in each instance (layout) we have a collection of solutions (turbines); this means that each data point is represented as a matrix $X_n \in \Rea^{|X|\times d}$. Secondly, we need to take into account the fact that layouts can have a different number of turbines. Therefore, we apply the Gaussian over sets as the Bayesian model and find a new set by maximising the acquisition function.

\subsection{Gaussian Processes Over Sets}
\label{section_gpos}
With $\mathcal{GP}s$ we aim to build a surrogate model that can predict the output of an objective function without an explicit evaluation. One of the major advantages of the Gaussian process as the Bayesian model is that it provides uncertainty in predictions. This uncertainty can help in finding the promising decision vectors in the subsequent iterations of the Bayesian optimisation.

A $\mathcal{GP}$ can be described as a multivariate normal distribution with mean $\boldsymbol{\mu}$ and covariance matrix $K$ \cite{Rasmussen2006} such that given a function $f$ we have:
$$
f \sim \mathcal{N}(\boldsymbol{\mu},K)
$$
The mean $\boldsymbol{\mu}$ and covariance matrix $K$ are both dependent on a kernel function which quantifies the correlation between instances in the data. For simplicity in calculations, we assume the zero mean. A variety of kernel functions can be used e.g.\ Radial basis function (RBF), Matern or linear kernel. In this work, we use a RBF (also known as squared exponential or Gaussian kernel): 
\begin{equation}
k(\mathbf{x}_1,\mathbf{x}_2,\boldsymbol{\Theta})= \sigma^2 \exp \Big(-  \frac{||\mathbf{x_1} -\mathbf{x_2}||^2}{2l^2} \Big) + \sigma_n^2 \delta_{\bx_1,\bx_2},
\label{ker_func}
\end{equation}
where $\boldsymbol{\Theta} = (l,\sigma,\sigma_n)$ is the vector of hyperparameters to be estimated when building the model, $||\mathbf{x_1} -\mathbf{x_2}||^2$ is the squared Euclidean distance between two decision vectors $\bx_1$ and $\bx_2$ and $\delta_{\bx_1,\bx_2}$ is the Kronecker delta function. The hyperparameters can be estimated by maximising the following likelihood function:
\begin{equation}
p(\mathbf{y}|\mathcal{D},\mathbf{\Theta}) = \frac{1}{\sqrt{|2\pi K|}} \mathrm{exp}(\mathbf{y}^T K ^{-1} \mathbf{y}),
\label{likelihood}
\end{equation}
where  $K_{i,j} =\: k(\mathbf{x}_i,\mathbf{x}_j,\mathbf{\Theta})$ with $\mathbf{x}_i,\mathbf{x}_j$ representing rows of the data matrix $X$ and $y = f(\bx)$. Once the optimal parameters are found, we can predict the value of the objective function of a new instance $\mathbf{x}'$ using the following posterior predictive distribution:

\begin{equation}
\begin{split}
p(\mathbf{y}'|\mathbf{x}',\mathcal{D},\mathbf{\Theta})  =  \mathcal{N} & (\mathbf{k}(\mathbf{x}',X)K^{-1}\mathbf{y},\\ & \:\: k(\mathbf{x}',\mathbf{x}') - \mathbf{k}(\mathbf{x}',X)K^{-1}\mathbf{k}(\mathbf{x}',X)),
\label{post_pred}
\end{split}
\end{equation}
where $\mathbf{k}(\mathbf{x}',X)$ is the covariance vector between $\mathbf{x}'$ and the training data $X$.
To deal with the problem of finding the correlation between sets, we utilise Gaussian Process Over Sets \cite{gpoversets} which is based on the idea of using a set of decision vectors. A $j^{th}$ design set can be written as:
\begin{equation}
    \begin{split}
    & X_j = [\mathbf{x}_1,\mathbf{x}_2,\ldots,\bx_n] \:\:\: \\
    & \mathrm{with} \:\:\: \mathbf{x}_i \in \Rea^{1\times d}
    \end{split}
\end{equation}
where $d$ is the number of decision variables and $n$ is the number of decision vectors. In wind farm layout optimisation, the x-coordinate and y-coordinate are the decision variables and therefore, $d=2$ and the number of turbines $n$ vary for different wind farms. The $\mathcal{GP}$ Over Sets requires adjustments to the traditional Gaussian Process algorithm, namely the way that the correlation between sets is calculated. For $\mathcal{GP}$ Over Sets, we compute correlation between sets as follows:
\begin{equation}
    \begin{split}
    k_{set}(X_1,X_2) &= \frac{1}{|X_1||X_2|} \sum_{\mathbf{x}_i \in X_1}\sum_{\mathbf{x}_j \in X_2} k(\mathbf{x}_i,\mathbf{x}_j)\\
    &\mathrm{where}\\
    & |X|\:\:\:\:\: \mathrm{Cardinality\: of \:set\:X}\\
    & k(\mathbf{x}_i,\mathbf{x}_j) \: \mathrm{is \:a \:kernel\: function \:(eg.\: equation (\ref{ker_func}))}
    \end{split}
\end{equation}
Two important advantages of using the kernel over sets mentioned above is the resulting covariance matrix is positive-definite and the order of the decision vectors in the set do not effect the correlations between sets. The data set for building the model with $N$ sets is:
\begin{equation}
    \begin{split}
        \mathcal{D} = \{\mathcal{X},Y\:|\mathcal{X} = [X_1,X_2,...X_N]\:\:\mathrm{with}\:\: X_i \in \Rea^{|X_i|\times d}, Y \in \Rea^{N\times w}\}
    \end{split}
\end{equation}
The posterior predictive distribution (equation (\ref{post_pred})) becomes:
\begin{equation}
\begin{split}
p(\mathbf{y}'|X',\mathcal{D},\mathbf{\Theta})  =  \mathcal{N} & (\mathbf{k}_{set}(X',\mathcal{X})K^{-1}\mathbf{y},\\ & \:\: k_{set}(X',X') - \mathbf{k}_{set}(X',\mathcal{X})K^{-1}\mathbf{k}_{set}(X',\mathcal{X})),
\end{split}
\end{equation}
where $K_{i,j} = k_{set}(X_i,X_j)$ and $\mathbf{k}_{set}(X',\mathcal{X})$ is the correlation vector between the new set $X'$ and the training data $\mathcal{X}$.
This formulation is compatible with the objective function formulation that is used in this paper since it allows to compute the correlation of instances with different lengths, where the length corresponds to the number of decision vectors in a set.

\subsection{Problem Encoding}
\label{section:problem_encoding}
\begin{figure}[t]
\centering
\includegraphics[scale=.5]{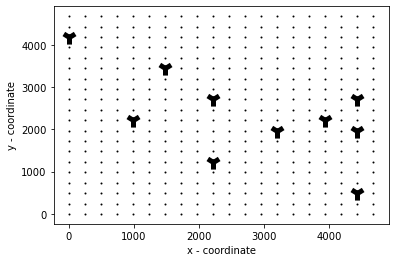}
\caption{Visual representation of the binary encoding. The $1s$ are the location of turbines and $0s$ are the empty space. The x and y coordinates are converted to $0s$ and $1s$ based on the location of turbines.}
\label{binary_encoding}
\end{figure}

In order to find a new design set, we use the expected hypervolume improvement as the acquisition function. The hypervolume is the $w$-dimensional Lebesgue measure of the space dominated by a finite approximation of the non-dominated solutions in the objective space $P = {\mathbf{f}^1,\ldots\mathbf{f}^n}$, which is bounded by a reference point $\mathbf{r}$:
\begin{align}
    HV(P) = \lambda_k(\cup_{\mathbf{f} \in P} [\mathbf{f},\mathbf{r}]).
\end{align}
The hypervolume improvement of a vector $\by$ is: 
\begin{align}
    HV(P,\mathbf{y}) = HV(P \cup \mathbf{y}) - HV(P)
\end{align}
and expected hypervolume improvement contribution of a new solution $\by$ is:
\begin{align}
    EHVI = \int_{\Rea^w} HV(P,\mathbf{y}) \cdot PDF_{\bm{\mu},\bm{\sigma}}(\mathbf{y}) d\mathbf{y},
\end{align}
where $\bm{\mu}$ and $\bm{\sigma}$ are the approximated mean and standard deviations from the Bayesian model. An optimiser e.g.\ an evolutionary algorithm can be used to maximise the EHVI. To have the same number of decision variables in maximising EHVI, we impose an encoding. We represent a wind farm with a finite number of grid points (where turbines can be installed). In this layout, the $1s$ represent a turbine being present in a specific grid point, and $0s$ represent an empty space in the grid. Note that, the Gaussian process as the Bayesian model is trained on two dimensional real coordinates, which are then encoded to 0 and 1 to maximise the acquisition function. An illustration of encoding representation of a wind farm (with $20 \times 20$ grid points) is shown in Figure \ref{binary_encoding}.
This work uses the binary genetic algorithm to maximise the EHVI in finding a new vector of 0s and 1s. This vector is then decoded back to real coordinates to get a design set. This design set is then used to calculate expensive objective functions and the model is re-trained. This encoding allows to easily switch between the real (for training the models) and binary coordinates (for maximising the acquisition function).


\section{Results and discussion}
We applied the Bayesian optimisation on a wind farm with maximum of 400 turbines spaced in $20 \times 20$ grid\footnote{We used the EHVI implementation available at\url{https://liacs.leidenuniv.nl/~csmoda/index.php?page=code}}. The rest of experimental settings are as follows:
\begin{enumerate}
    \item Initial number of sets (or wind farms) = 20 (as $10 \times d$), generated randomly with minimum of two and maximum of 400 turbines
    \item Kernel function in Gaussian process: Radial basis (or squared exponential) with noise
    \item Algorithm to maximise the likelihood function: Differential evolution from SciPy \cite{scipy}
    \item Algorithm to maximise the EHVI as the acquisition function: Binary Genetic algorithm from PyMOO \cite{pymoo} 
    \item Maximum number of function evaluations: 100 including 20 initial sets
    \item Maximum number of independent runs = 3 
    \item Diameter of rotor of the turbine: 82 meters (standard diameter of a turbine)
    \item Data set to find the joint distribution of wind speed and wind direction: Engie Open Data (an open source data set) \cite{mydata}
    \item Method to find the joint distribution between wind speed and direction: kernel density estimation from Scipy with bandwidth selection as per the Scott's rule \cite{Scott1992}
    \item Minimum distance between turbines = $3 \times \text{diameter of rotor}$
    \item Wake model: Jensen from the PyWake library \cite{pywake}
\end{enumerate}

\begin{figure}
\centerline{\includegraphics[scale=.3]{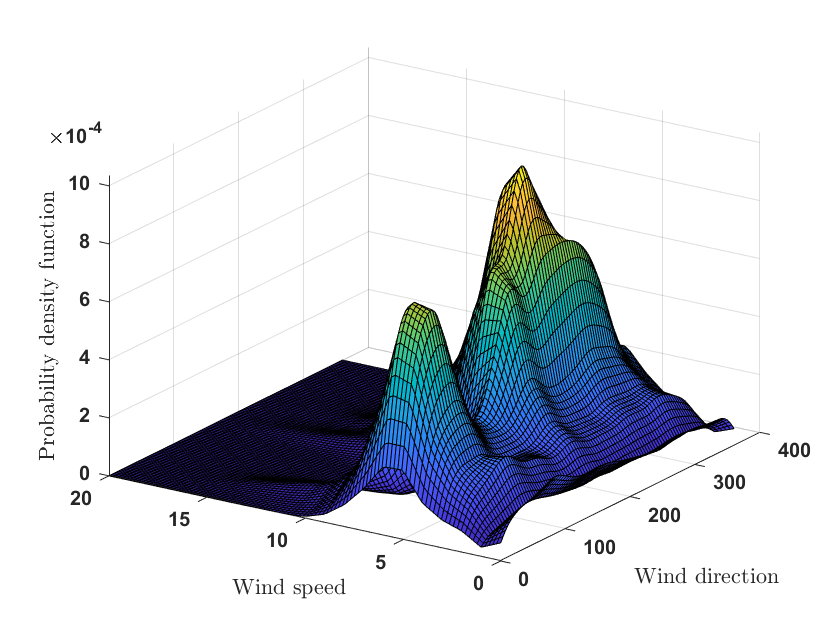}}
\caption{Probability density function of the joint distribution of wind speed (m/sec) and wind direction.}
\label{wind_pdf}
\end{figure}

\begin{figure}
\centerline{\includegraphics[scale=.3]{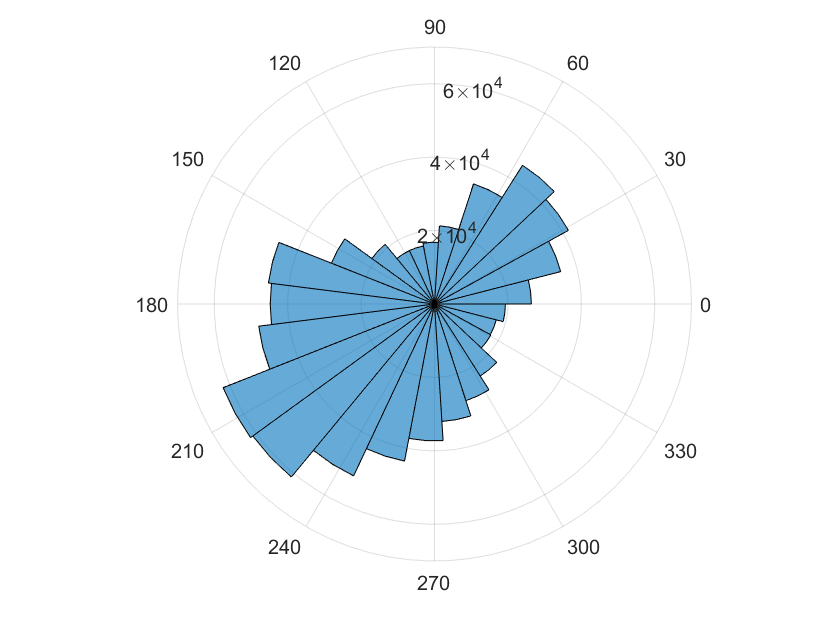}}
\caption{Polar histogram of wind direction}
\label{fig:polar_plot}
\end{figure}

The joint distribution of wind speed and direction estimated with kernel density estimation is shown in Figure \ref{wind_pdf}. As can be seen, there are two major peaks in the distribution, which are at the Southwest and Northeast directions as also can be seen in the polar histogram plots of wind direction in Figure \ref{fig:polar_plot}. 

\begin{figure}
\centerline{\includegraphics[scale=.5]{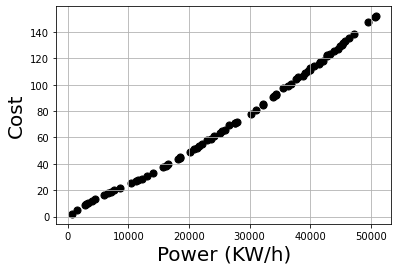}}
\caption{Approximated Pareto front - Power is to be maximised and cost is to minimised. The solutions are from the run with the median hypervolume.}
\label{pareto_fig}
\end{figure}

The joint distribution of wind speed and direction and wind speed estimations facing the turbines via wake model were used in getting the power of the wind farm. After running the algorithm for 100 expensive evaluations, we obtained an approximated Pareto front and is shown in Figure \ref{pareto_fig} (for the run with median hypervolume value). As can be seen, the algorithm was able to explore in different regions of the Pareto front and obtained a set of trade-off solutions. 

\begin{figure*}
 \begin{subfigure}[b]{0.3\textwidth}
   \includegraphics[scale=0.35]{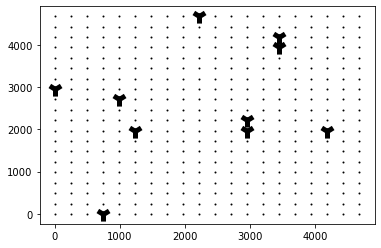}
   \caption{}
   \label{lay:K_1} 
\end{subfigure}
 \begin{subfigure}[b]{0.3\textwidth}
   \includegraphics[scale=0.35]{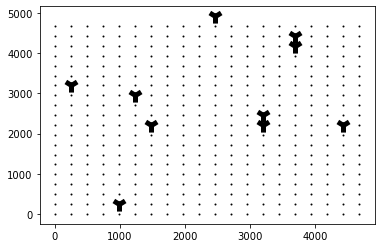}
   \caption{}
   \label{lay:K_2} 
\end{subfigure}
 \begin{subfigure}[b]{0.3\textwidth}
   \includegraphics[scale=0.35]{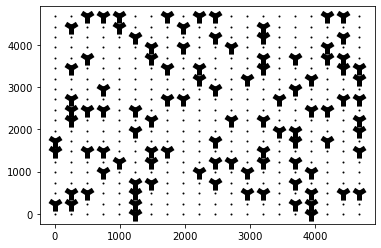}
   \caption{}
   \label{lay:K_3} 
\end{subfigure}

\begin{subfigure}[b]{0.3\textwidth}
   \includegraphics[scale=0.35]{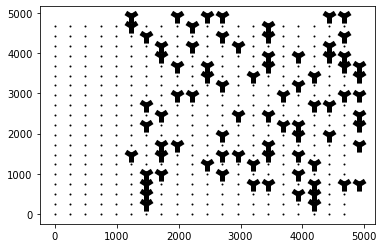}
   \caption{}
   \label{lay:K_4} 
\end{subfigure}
\begin{subfigure}[b]{0.3\textwidth}
   \includegraphics[scale=0.35]{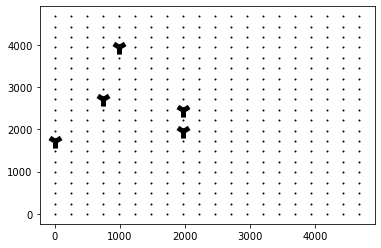}
   \caption{}
   \label{lay:K_5} 
\end{subfigure}
\begin{subfigure}[b]{0.3\textwidth}
   \includegraphics[scale=0.35]{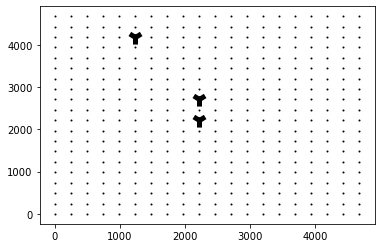}
   \caption{}
   \label{lay:K_6} 
\end{subfigure}
    
    \caption{Six wind farms for the illustration of the correlation between wind farms }
    \label{fig:wind_farms_K}
\end{figure*}

\begin{figure}
    \centering
    \includegraphics[scale=0.3]{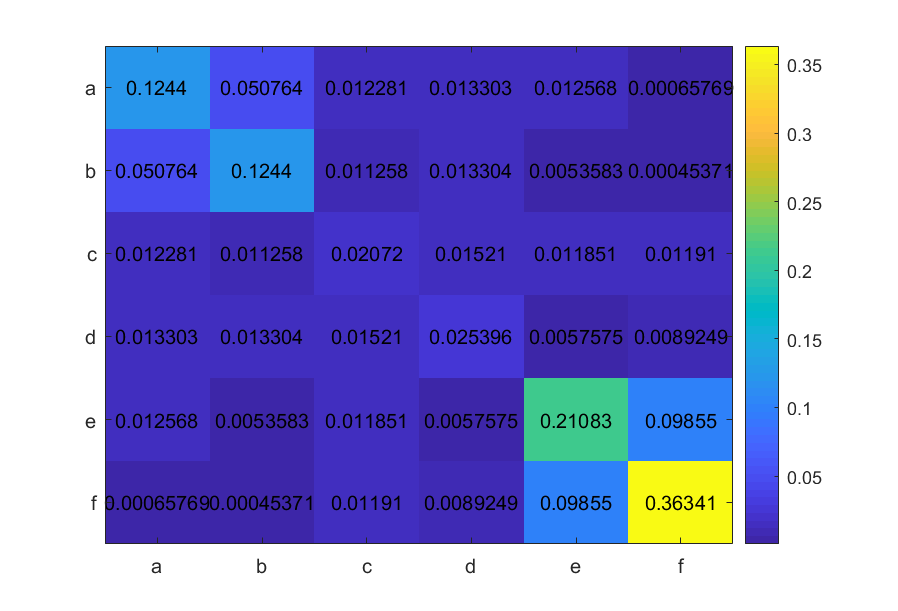}
    \caption{Correlation matrix with length scale, $l = (82 \times 3)$, amplitude $\sigma^2=1$, and noise variance $\sigma_n^2=1$ between six wind farms shown in Figure \ref{fig:wind_farms_K}.}
    \label{fig:K_matrix}
\end{figure}

Each point on the approximated Pareto front represents a wind farm with a different number of turbines. The set based kernel in Gaussian process consider the correlation between wind farms. In other words, two wind farms with similar x and y coordinates have a large correlation compared to two wind farms with different x and y coordinates. For instance, consider six wind farms $(a, b, c, d, e, f)$ shown in Figure \ref{fig:wind_farms_K}. These wind farms have different number of turbines. Some of the wind farms e.g.\ (`a' and `b'), (`c' and `d') and (`e' and `f') have turbines in the similar locations. Therefore, we expect to see a higher correlation between these pairs compared to e.g.\ (`a' and `e') or (`b' and `d'). For a length scale, $l = (82 \times 3)$, amplitude $\sigma^2=1$, and noise variance $\sigma_n^2=1$, we plot the covariance matrix representing the correlation between these wind farms in Figure \ref{fig:K_matrix}. As can been seen, obviously the correlation values are the largest among diagonal elements. After the diagonal elements, the correlation values follow the kernel calculations. For example, the wind farms in the decreasing order of correlation with wind farm `a' are (`a', `b', `d', `e', `c', `f') and with  wind farm `e' are (`e', `f', `a', `c', `d', `b'). 

\begin{figure}[h]
\centering
\includegraphics[scale=.5]{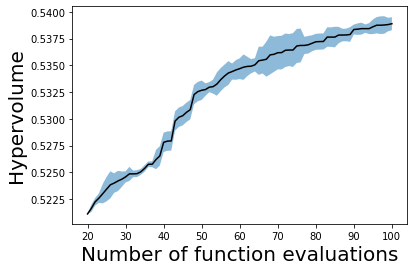}
\caption{Hypervolume with the number of expensive evaluations}
\label{hv_changes}
\end{figure}

\begin{figure*}
\begin{subfigure}[b]{0.3\textwidth}
   \includegraphics[width=1\linewidth]{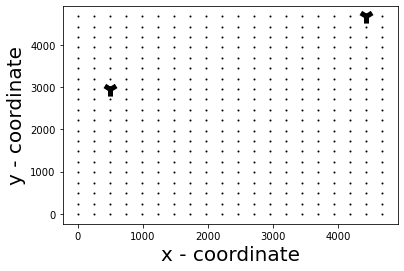}
   \label{lay:1} 
\end{subfigure}
\begin{subfigure}[b]{0.3\textwidth}
   \includegraphics[width=1\linewidth]{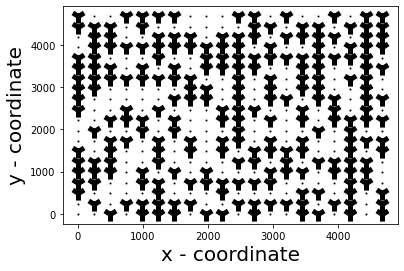}
   \label{lay:2} 
\end{subfigure}
\begin{subfigure}[b]{0.3\textwidth}
   \includegraphics[width=1\linewidth]{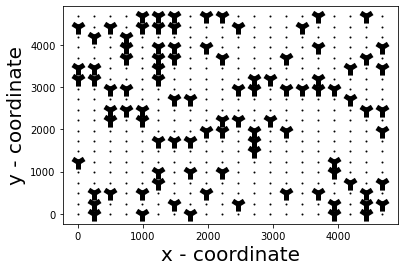}
   \label{lay:3} 
\end{subfigure}
\caption{Three randomly selected wind farms after running the algorithm. The power and cost objective function values for the top wind farm are (632.69, 1.99), for the middle wind farm are (5.08 $\times 10^4$, 152) and for the bottom wind farm are (2.54 $\times 10^4$, 64.67).}
\label{fig:lay}
\end{figure*}

Figure \ref{hv_changes} shows the hypervolume with the number of function evaluations. The solid line is the median and shaded region represents the 95\% confidence interval. As can be seen, the hypervolume increased with the number of function evaluations. 
Figure \ref{fig:lay} shows three wind farms (or sets) from the approximated Pareto front. The first two wind farms are the extreme points on the approximated Pareto front with 2 and 228 turbines. The third wind farm has 97 turbines. We did not select one wind farm as the final solution as it depends on the expert or decision maker to select a wind farm based on the objective function values and their preferences. Presenting wind farms with their objective function values can provide useful insights to decision makers to make an informed decision for wind farm layout design.

\section{Conclusions}
mIn this work, we applied Multi-objective Bayesian optimisation with expected hypervolume improvement as the acquisition function to handle the computationally expensive optimisation problem. To handle the uncertainty in wind direction and wind speed, we modelled the data with kernel density estimation and used their joint distribution. As the power output of the wind farm relied on design sets, we utilised the kernel over sets in Gaussian process and embedded it into the multi-objective Bayesian optimisation framework. As shown in the results, the method was able to obtain a set of approximated Pareto optimal solutions. The method was able to handle and quantify the correlation between different wind farms with different number of turbines. The hypervolume as the performance metric showed the validity of the method. 

This paper presented the first attempt at trying to solve the wind farm layout optimisation using Multi-objective Bayesian optimisation over sets. We believe that there are particular aspects of the model that can be further investigated and improved. For example, finding correlations between wind farms could be improved with a different set based kernel. Moreover, the computational complexity of set based kernel is $O(N^2 |X|^2 d)$, which is higher than the computational complexity $O(N^3)$ of traditional Gaussian process, ($N$ is the number of sets (or vectors in traditional Gaussian process), $|X|$ is the cardinality of the set and $d$ is the number of decision variables. Further research could provide ways of reducing the computational cost of the model. In this work, we built a surrogate model for the cost function which was not computationally expensive and was a function of the number of turbines. This modelling of the cost function may not be required to maximise the expected hypervolume improvement. Working on such heterogeneous objective functions in Bayesian optimisation is also a future research direction. Utilising different wake models including computational fluid dynamic (CFD) solvers will also be considered as one of the main future research directions. We believe that this work can lead to an increase interest in set based optimisation especially in Multi-objective Bayesian optimisation which will widen its application to other research areas, especially where the objective functions rely on a set of solutions.

\section*{Acknowledgments}
The authors would like to thank Prof.\ Kishalay Mitra, Prof.\ Jonathan Fieldsend, Prof.\ Richard Eversion, Dr Srinivas Soumitri Miriyala and Dr Priyanka D Pantula for initial discussions on optimising wind farms. This research was partially supported by South Asia Development Fund at the University of Exeter, UK.

\bibliographystyle{plain}
\bibliography{main}

\end{document}